%% file: main.tex
\documentclass[conference, letterpaper]{IEEEtran}
\IEEEoverridecommandlockouts

\usepackage{graphicx}
\usepackage{xcolor}
\usepackage{amsmath, amssymb, amsfonts}
\usepackage{booktabs}
\usepackage{siunitx}
\usepackage[nolist,nohyperlinks]{acronym}
\usepackage{url}
\usepackage{mathtools}
\usepackage{bm}
\usepackage{subcaption}
\usepackage[implicit=false]{hyperref}
\usepackage{tabularx}
\usepackage{multirow}
\usepackage{makecell}
\usepackage{float}
\usepackage{pifont}
\usepackage{arydshln}
\usepackage{csquotes}
\usepackage{censor}
\usepackage{cite}
\usepackage{todonotes}

\usepackage[placement=top,vshift=-25]{background}
\newcommand{\hdrvenue}{IEEE/RSJ IROS 2026. PREPRINT VERSION. ACCEPTED JUNE, 2026}
\newcommand{\hdrtitle}{PLISKA \textit{et al.}: TOWARDS AGILE VISION-BASED MULTI-UAV FLIGHT: REVISITING STATE ESTIMATION}
\newcommand{\runninghead}{\footnotesize\makebox[\textwidth]{%
  \ifnum\value{page}=1 \hdrvenue\hfill\thepage
  \else\ifodd\value{page}\hdrtitle\hfill\thepage
  \else\thepage\hfill\hdrvenue\fi\fi}}
\SetBgScale{1.0}
\SetBgContents{\runninghead}
\SetBgColor{black}
\SetBgAngle{0}
\SetBgOpacity{1.0}

\input{acronyms.tex}

\input{common.tex}

\title{\vspace*{10pt}Towards Agile Vision-Based Multi-UAV Flight: Revisiting State Estimation}

\author{Michal Pliska, Matouš Vrba, Ondřej Víta, Martin Jiroušek, Viktor Walter and Martin Saska%
\thanks{Authors are with the Multi-robot Systems Group, Faculty of Electrical Engineering, Czech Technical University in Prague, Czech Republic (\texttt{name.surname@fel.cvut.cz}). This work was funded by the Czech Science Foundation (GA\v{C}R) under projects no.~26-22419S and 23-06162M.}%
}

\begin{document}

\maketitle

\begin{abstract}
Agile multi-\acs{uav} flight requires accurate and low-latency onboard estimation of the kinematic states of neighboring \acsp{uav} for collision avoidance, motion coordination, etc.
Most vision-based approaches rely on position-only measurements, inferring velocity and acceleration indirectly from displacement.
We show that this introduces a fixed structural delay in the estimation of higher-order states, which limits the achievable agility.
To address this, we propose to integrate tilt measurements, provided by a state-of-the-art visual detector, which inform about the thrust direction of co-planar multirotor \acsp{uav}.
We benchmark four position-only and five pose-aware estimators, including a novel formulation of a linear thrust-constraining \acl{kf}, on two real-world and one high-fidelity photorealistic simulated dataset over different levels of agility (\qtyrange{3}{21}{\metre\per\second\squared}).
In our setup, pose-aware estimation consistently reduces the average velocity and acceleration estimation errors by \SI{40}{\percent} and \SI{57}{\percent} across the three datasets with the proposed \acs{kf} formulation outperforming the other estimators.
Position-only filters exhibit a constant ${\sim}\SI{300}{\milli\second}$ delay in acceleration step response independent of agility, whereas the tilt-constrained estimators operate near the physical response limit given by the camera frame-rate by observing the change in thrust direction before the displacement accumulates.
In a closed-loop leader--follower simulated experiment with NMPC control, position-only estimation of the leader's state fails to facilitate stable hovering of the follower, while the proposed estimator enables tracking of lateral maneuvers exceeding \SI{2}{\gee} of acceleration.
\end{abstract}


\vspace{-0.7em}
\section*{Supplementary Material}
{\small
\vspace{-0.3em}
\noindent \textbf{Video, code \& data:} \url{https://mrs.fel.cvut.cz/agile-uav-estimation}
\vspace{-0.5em}}

\section{Introduction}

\global\csname @topnum\endcsname 0
\global\csname @botnum\endcsname 0

\begin{figure}[t]
  \centering
  \captionsetup{skip=6pt, labelfont=bf, labelsep=period, font=small}
  \includegraphics[width=\columnwidth,clip,trim=0 0mm 0 0mm]
    {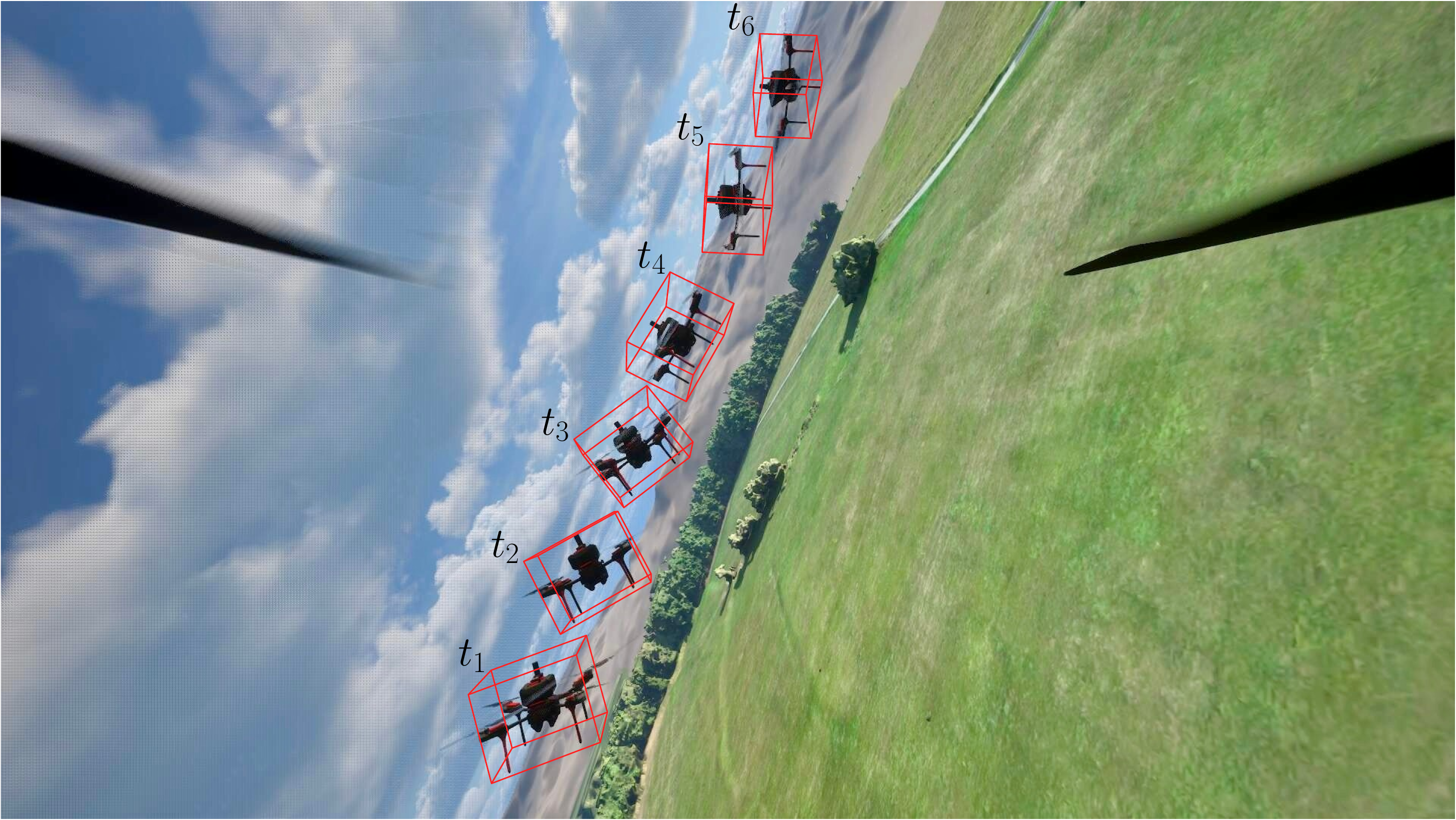}
\caption{%
Onboard view from a follower UAV during closed-loop tracking at accelerations above \SI{2}{\gee}.
Red boxes show raw YOLOv5-6D detections.
The first frame ($t_1$) is the current image, the remaining frames ($t_2$--$t_6$) are a time-lapse during an aggressive lateral maneuver.
The times $t_1$--$t_6$ correspond to the markers in Fig.~\ref{fig:closedloop}.%
}
\label{fig:sim_timelapse}
\end{figure}

Recent advancements in multirotor \ac{uav} control have pushed the agility of individual autonomous \acp{uav} to the edge of the platforms' capabilities, surpassing even expert human pilots~\cite{guptaLoLNMPCLowLevelDynamics2025, song2023reaching, feredeOneNetRule2025}.
However, state-of-the-art multi-robot \ac{uav} systems lag significantly behind these advances in agility.
This is especially true for multi-\ac{uav} systems that rely on onboard relative localization to remove the requirement of intra-robot communication, which introduces non-negligible measurement noise~\cite{opromollaVisionBasedApproachUAV2018, walterUVDARSystemVisual2019}.
While the problem has been extensively studied~\cite{schillingVisionBasedDroneFlocking2021, tangVisionAidedMultiUAVAutonomous2019, zhangAgileFormationControl2022}, the reported experiments are typically conducted at velocities and accelerations that remain well below the agility routinely achieved in single-\ac{uav} scenarios with full state knowledge.
In this work, we present the building blocks necessary to fill this research gap.

When considering multirotor \ac{uav} dynamics, direction of the thrust vector provides useful information about an observed target's state and is often directly measurable, whether through the vehicle's attitude, propeller orientation, or other visual cues.
For a typical multirotor platform with co-planar propellers, the thrust vector is aligned with the direction of the propellers.
There are several works that propose methods to exploit this feature of multirotor dynamics using visual detection of the full 6D pose to measure the \ac{uav}'s position and attitude~\cite{jinDroneDetectionPose2019a, zhengKeypointGuidedEfficientPose2024a, 11373013}.
Despite this growing recognition, it remains unknown to what extent including the thrust direction measurement improves state estimation over position-only methods.
In particular, no systematic comparison across agility regimes with consistent tuning procedures and explicit finite-time latency analysis exists.

This paper addresses this gap and presents the first systematic comparison of pose-aware and position-only multirotor \ac{uav} state estimation, conducted over a full monocular perception-to-estimation pipeline across real-world and simulated datasets spanning multiple agility levels.
The best position-only and pose-aware methods are also compared in an agile leader-follower scenario running the whole pipeline from \ac{uav} control, onboard visual detection and state estimation of the leader, to the formation control (see Fig.~\ref{fig:sim_timelapse}).
These experiments demonstrate that the pose-aware relative state estimation is a necessary element for realizing truly agile multi-\ac{uav} motion coordination approaching the dynamic limits of individual \acp{uav}.

\section{Related Work}
\label{sec:related}

\subsection{Onboard \acs{uav} detection}

Classical approaches of onboard detection of other \acp{uav} rely on various kinds of markers, such as LEDs placed onto the targets to be detected and localized~\cite{walterUVDARSystemVisual2019}, which enables reliably extracting their full poses without relying on domain-specific training.
A distinct subset of \ac{uav} detection methods relies on sensors such as stereo cameras or LiDARs to detect flying \acp{uav} using the spatial information provided by these sensors~\cite{carrioOnboardDetectionLocalization2020a, vrbaOnboardLiDARBasedFlying2025}.
This approach inherently yields the relative 3D position of the target, but requires equipping the observer \ac{uav} with specialized, potentially heavy, and expensive sensors.

Recently, deep learning-based methods became sufficiently robust, accurate, and computationally efficient for deployment onboard \acp{uav} with limited payload~\cite{vrbaMarkerLessMicroAerial2020, barisicBrainBrawnUsing2022}.
Modern deep-learning detectors can extract additional information about the target in addition to its relative 3D position, such as the full 6D pose, using cheap and lightweight monocular RGB cameras and without relying on markers~\cite{zhengKeypointGuidedEfficientPose2024a, jinDroneDetectionPose2019a, 11373013}.
At the propeller level, Špetlík et al.~\cite{Spetlik_2026_CVPR} proposed a method for real-time roll and pitch estimation from event-camera streams.
Although similar event-based approaches have so far only been demonstrated in controlled laboratory conditions, they have the potential to offer a complementary sensing modality to RGB-based pose estimation.

In this work, we utilize the YOLOv5-6D deep learning architecture~\cite{viviersAdvancing6DoFInstrument2024} that relies on visual keypoint detection, and \ac{pnp} to obtain full 6D poses of the detected objects assuming their known dimensions.

\subsection{\ac{uav} tracking and state estimation}

State estimation of aerial targets is most often formulated through position-based \acp{kf} using simple kinematic models, most commonly \acf{cv} or \acf{ca}~\cite{vrbaMarkerLessMicroAerial2020, barisicBrainBrawnUsing2022, vrbaOnboardLiDARBasedFlying2025, dogruDroneDetectionUsing2022}.
These models assume smooth motion and compensate for any unmodeled dynamics with process noise, which is typically tuned empirically.
While sufficient for moderately maneuvering targets (max. acceleration reported in~\cite{vrbaMarkerLessMicroAerial2020, barisicBrainBrawnUsing2022, vrbaOnboardLiDARBasedFlying2025, dogruDroneDetectionUsing2022} was \SI{4.5}{\metre\per\second\squared}), this approach introduces a fundamental trade-off: low process noise yields estimation lag, whereas high noise produces jitter and instability~\cite{pliskaSafeMidAirDrone2024a}.
Crucially, since these approaches rely exclusively on position measurements, uncertainty propagates and amplifies into higher-order state estimates.

Several recent works have begun to incorporate attitude information into \ac{uav} state estimation.
The authors of~\cite{jinDroneDetectionPose2019a} proposed a custom convolutional neural network for extracting salient keypoints of a co-planar \ac{uav} combined with the standard \ac{pnp} algorithm for full pose estimation, and a \ac{ca}-\ac{kf} relying on an assumption of zero vertical acceleration of the observed target.
A very similar approach was described in~\cite{zhengKeypointGuidedEfficientPose2024a}, but with a more general \ac{plkf}-based estimator.
\cite{zhengKeypointGuidedEfficientPose2024a} also publishes a dataset for \ac{uav} pose detection obtained in laboratory conditions with motion capture ground truth, which we utilize in this work for baseline comparison.
A complementary approach is presented in~\cite{11373013}, removing the assumption of known physical dimensions of the tracked target, and providing an observability analysis of the problem.

However, all these works evaluate their methods on scenarios with limited dynamic excitation of the target and negligible vertical motion, which does not sufficiently cover the intended application of agile multi-robot motion.
These works also do not report the method used for tuning the estimation parameters.

\subsection{Contributions}

The presented paper provides a systematic comparison of position-only and pose-aware vision-based state estimation of co-planar multirotor \acp{uav}.
We compare performance across different agility levels using identical automated tuning for all estimators to ensure fairness.
To facilitate this, we release a new \ac{uav} dataset created in high-fidelity simulations containing diverse maneuvers of varying agility.
Furthermore, we propose the \ac{dkf} framework for formulating a linear \ac{kf} with general dynamic affine subspace measurements of the state vector, which is a common problem in robotics~\cite{zhengKeypointGuidedEfficientPose2024a, 11373013, mourikis2007multi}.
Using this framework, we derive the \ac{zlkf} -- an efficient linear \ac{kf} variant for the task of pose-aware \ac{uav} state estimation, which outperforms other state-of-the-art methods.
Finally, we demonstrate a simulated experiment running the full pipeline, including \ac{uav} control, onboard perception, state estimation, and trajectory prediction of the target in a leader-follower scenario with high agility (leader reaching velocities of \SI{14}{\metre\per\second} and accelerations of \SI{20}{\metre\per\second\squared}).
We believe that our work will serve as a springboard for other researchers aiming to push their multi-\ac{uav} systems from idealized simulations and laboratories to challenging real-world conditions.

In the context of the state of the art, we summarize our contributions as follows:
\begin{enumerate}
  \item The first systematic comparison of pose-aware and position-only state estimation for vision-based \ac{uav} tracking is provided, evaluated through a full monocular perception-to-estimation pipeline over different levels of agility w.r.t. the target's acceleration.

  \item A novel linear \ac{kf} formulation that incorporates pose as a directional acceleration constraint for improved \ac{uav} state estimation is formally derived using the \acf{dkf} framework.

  \item A benchmark dataset with photorealistic rendering and exact ground truth across seven agility levels is released, enabling reproducible comparison of estimators for agile full-pose vision-based \ac{uav} state estimation.

  \item The impact of the state estimation on communication-free \ac{uav} motion coordination is evaluated in a simulated closed-loop leader-follower scenario, demonstrating that the proposed pose-aware estimation enables reaching unprecedented agile levels of multi-\ac{uav} flight.
\end{enumerate}

\section{State Estimation}

\begin{figure}[t]
  \centering
  \captionsetup{skip=6pt, labelfont=bf, labelsep=period, font=small}
  \includegraphics[width=0.7\columnwidth]
    {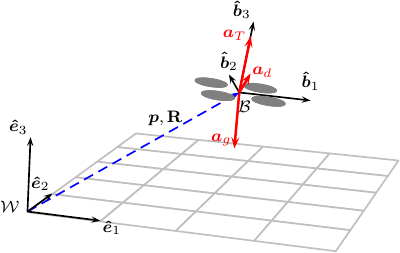}
  \caption{Relation of the world frame $\mathcal{W}$ and the \ac{uav}'s body frame $\mathcal{B}$ through the position vector $\vec{p}$ and orientation matrix $\mat{R}$. Basis vectors of these frames and the relevant acceleration components of the total \ac{uav} acceleration $\vec{a} = \vec{a}_T + \vec{a}_d + \vec{a}_g$ are also denoted.}
  \label{fig:coordinate_frames}
\end{figure}

We model the full \ac{uav} dynamics using a set of non-linear differential equations
\begin{align}
  \ddot{\vec{p}} &= \frac{1}{m} \left( f_T \mat{R} \vec{\hat{e}}_3 + \vec{f}_d \right) - g\vec{\hat{e}}_3, \label{eq:main_model}\\
  \dot{\mat{R}} &= \mat{R} \mat{\Omega},
\end{align}
where $\vec{p} \in \mathbb{R}^3$ is the position of the \ac{uav}, $m$ is its mass, $f_T$ is its collective thrust force, $\mat{R} \in \mathbf{SO}(3)$ represents its orientation, $\mat{\Omega} \in \mathbb{R}^{3\times 3}$ is an angular velocity tensor expressing the angular velocity of the \ac{uav},
$\vec{\hat{e}}_3 = \bemat{0 & 0 & 1}\tran$ is the third canonical basis vector, $\vec{f}_d$ is an external disturbance force, and $g$ is the gravitational acceleration constant.

Furthermore, we denote the \ac{uav}'s body z-axis vector as $\vec{\hat{b}}_3$, its velocity vector as  $\vec{v} \in \mathbb{R}^3$, its acceleration vector as $\vec{a} \in \mathbb{R}^3$, its jerk vector as $\vec{j} \in \mathbb{R}^3$, the acceleration component corresponding to the thrust as $\vec{a}_T = \frac{1}{m} f_T \mat{R} \vec{\hat{e}}_3$, to the external disturbance as $\vec{a}_d = \frac{1}{m} \vec{f}_d$, and to the gravity as $\vec{a}_g = -g \vec{\hat{e}}_3$ (see Fig.~\ref{fig:coordinate_frames}).
Where relevant, we denote a time-varying variable at the $k$-th time step with a lower index $(\cdot)\tstep{k}$.
Unless otherwise noted, all variables are expressed in the inertial world frame $\mathcal{W}$.
The state estimators described in this section are all based on this model with varying degrees of simplification.

\subsection{Point Mass-Model \acl{kf}}
\label{sec:mp_lkf}
As a baseline solution, we use a \ac{kf} with only position measurements and a point mass model.
This is a widely used approximation of the \ac{uav} dynamics when only position measurements are available and represents most of the SotA approaches~\cite{vrbaMarkerLessMicroAerial2020, barisicBrainBrawnUsing2022, vrbaOnboardLiDARBasedFlying2025, dogruDroneDetectionUsing2022}.
We follow a common formulation of a stochastic dynamic system within the \ac{kf} framework
\begin{align}
  \vec{x}\tstep{k+1} &= \mat{A} \vec{x}\tstep{k} + \vec{\xi}\tstep{k}, && \vec{\xi}\tstep{k} \sim \mathcal{N}\left( \vec{0}, \mat{\Xi} \right), \\
  \vec{z}\tstep{k} &= \mat{H} \vec{x}\tstep{k} + \vec{\zeta}\tstep{k}, && \vec{\zeta}\tstep{k} \sim \mathcal{N}\left( \vec{0}, \mat{Z} \right). \label{eq:meas_model}
\end{align}
Here, 
$\vec{x} \in \mathbb{R}^{n_x}$ is the state vector, $\vec{\xi} \in \mathbb{R}^{n_x}$ is the process noise with a covariance $\mat{\Xi}$, $\mat{A} \in \mathbb{R}^{n_x\times n_x}$ is the state transition matrix,
$\vec{z} \in \mathbb{R}^{n_z}$ is the measurement vector, $\vec{\zeta} \in \mathbb{R}^{n_z}$ is the measurement noise with a covariance $\mat{Z}$, $\mat{H} \in \mathbb{R}^{n_z\times n_x}$ is the measurement matrix, $\vec{0}$ is a $3$-dimensional zero vector, and $n_x$, $n_z$ are dimensions of the state-space and measurement-space.

We consider the \acf{cv} and \acf{ca} variants of the point mass-model \ac{kf} with the corresponding definitions of $\vec{x}$ and $\mat{A}$.
In both cases, the position measurement is modeled as
\begin{align}
  \vec{z}\tstep{k} = \vec{z}\tstep[p]{k} &= \vec{p}\tstep{k} + \vec{\zeta}\tstep[p]{k}, && \vec{\zeta}\tstep[p]{k} \sim \mathcal{N}\left( \vec{0}, \sigma_p^2\mat{I} \right),
\end{align}
where $\vec{z}\tstep[p]{k}$ is the position measurement, $\sigma_p$ is its per-axis standard deviation, and $\mat{I}$ is a $3\times 3$ identity matrix.
For both variants of the model, two sub-variants are considered: one with an \textit{analytical covariance matrix} of the process noise, and the second with a \textit{block-diagonal covariance matrix} of the process noise (abbreviated BDC).

\subsubsection{Analytical covariance of the process noise}
For the first sub-variant of the point mass-based \acp{kf}, the process noise has a physical interpretation as the unknown input of the system, which is assumed to follow a Gaussian distribution:
\begin{align}
  \vec{\xi}\tstep{k} &= \mat{B} \vec{u}\tstep{k}, && \mat{\Xi} = \mat{\Xi}_{\text{ana.}} = \mat{B} \mat{\Xi}_u \mat{B}\tran,
\end{align}
where the input $\vec{u}$ corresponds to acceleration for the \ac{cv} variant or to jerk for the \ac{ca} variant, and $\mat{\Xi}_u \equiv \sigma_u^2 \mat{I}$.

\subsubsection{Block-diagonal covariance of the process noise}
The second sub-variant assumes possible cross-correlation of the position, velocity and acceleration states, but no cross-correlation between their elements.
This formulation is more general and can partially mitigate model approximations and unmodeled dynamics, but it is defined by more parameters and is more prone to divergence during tuning.
We define the BDC matrix for the \ac{ca} variant as
\begin{align}
  \mat{\Xi} = \mat{\Xi}_{\text{BDC}} = \left[\begin{smallmatrix}
    \sigma_{p}^2 & \sigma_{pv}^2 & \sigma_{pa}^2 \\
    \sigma_{pv}^2 & \sigma_{v}^2 & \sigma_{va}^2 \\
    \sigma_{pa}^2 & \sigma_{va}^2 & \sigma_{a}^2
  \end{smallmatrix}\right] \otimes \mat{I},
\end{align}
and accordingly for the \ac{cv} variant.
Here, $\sigma_{(\cdot)}$ are parameters of this matrix, and $\otimes$ denotes the Kronecker product.
In other words, the BDC matrix is symmetrical and consists of $3\times 3$ diagonal sub-matrix blocks.

\subsection{Z-Axis Measurement \acl{kf} (\acs{zlkf})}
\label{sec:zlkf}

In this section, we derive a linear estimator based on the \ac{ca}-\ac{kf} described in the previous section adapted for fusing the measured tilt of the target \ac{uav}.
This is achieved by representing the tilt measurement as the z-axis vector $\hat{\vec{b}}_3$ of the \ac{uav}'s body frame $\mathcal{B}$ (i.e. the upward-pointing axis), which constrains the direction of the thrust acceleration $\vec{a}_T$.
By substituting for accelerations in eq.~\eqref{eq:main_model}, the total \ac{uav}'s acceleration is expressed as
\begin{equation}
  \vec{a}\tstep{k} = \vec{a}\tstep[T]{k} + \vec{a}_g + \vec{a}\tstep[d]{k} = \frac{f\tstep[T]{k}}{m}\vec{\hat{b}}\tstep[3]{k} + \vec{a}_g + \vec{a}\tstep[d]{k}, \label{eq:acc_model}
\end{equation}
where $\vec{\hat{b}}\tstep[3]{k} = \mat{R}\tstep{k}\vec{\hat{e}}_3$ is the z-axis of the \ac{uav} expressed in the world frame.
It is worth noting that we ignore the heading of the \ac{uav} (and thus the vectors $\hat{\vec{b}}_1$ and $\hat{\vec{b}}_2$) since it does not contribute to the \ac{uav}'s acceleration.
The modified measurement model is then
\begin{align}
  \vec{z}\tstep{k} &= \bemat{\vec{z}\tstep[p]{k}\tran & \vec{z}\tstep[\hat{b}_3]{k}\tran}\tran, \\
  \vec{z}\tstep[\hat{b}_3]{k} &= \vec{\hat{b}}\tstep[3]{k} + \vec{\zeta}\tstep[\hat{b}_3]{k}, && \vec{\zeta}\tstep[\hat{b}_3]{k} \sim \mathcal{N}\left( \vec{0}, \sigma_{\hat{b}_3}^2\mat{I} \right),
\end{align}
where $ \vec{z}\tstep[\hat{b}_3]{k}$ is the tilt measurement, and $\sigma_{\hat{b}_3}$ is the corresponding standard deviation.
The position measurement is expressed trivially within the \ac{kf} framework, but the z-axis vector is more involved.
Since the acceleration magnitude $\frac{f\tstep[T]{k}}{m}$ as defined in eq.~\eqref{eq:acc_model} is unknown, the actual acceleration $\vec{a}\tstep{k}$ lies within the acceleration space $\mathbb{R}^3$ on a line with direction $\vec{\hat{b}}\tstep[3]{k}$ and offset by $\vec{a}\tstep[g]{k} + \vec{a}\tstep[d]{k}$ from the origin.
Such line is an affine subspace of the acceleration space, which is different for every measurement.
To fuse this type of measurements, we introduce the \ac{dkf} framework.

\subsubsection{\acf{dkf}}
The \ac{dkf} framework defines a projection of the state vector similar to a \ac{kf} with state equality constraints~\cite{simon2006optimal}.
It may be considered a general step-by-step recipe to formulate a \acf{plkf} for a system with \enquote{degenerate} measurements that represent affine subspaces of the state space.
For brevity, we will omit the time indices in this subsection.

First, let us define a matrix $\mat{M} \in \mathbb{R}^{n_d \times n_x}$ that maps the full state space $\mathbb{R}^{n_x}$ to a subspace of the observed states $\mathbb{R}^{n_d}$ (i.e. the position and acceleration in the case discussed above).
Next, let the matrix $\mat{W} \in \mathbb{R}^{n_d \times n_a}$ be the basis of the measurement affine subspace of dimension $n_a$, with an unknown scaling vector $\vec{\lambda} \in \mathbb{R}^{n_a}$, and $\vec{o} \in \mathbb{R}^{n_d}$ is its offset vector.
The affine subspace measurement is then defined as
\begin{align}
  \mat{M}\vec{x} = \mat{W}\vec{\lambda} + \vec{o} + \vec{\zeta}', && \vec{\zeta}' \sim \mathcal{N}\left( \vec{0}, \mat{Z}' \right), \label{eq:dkf_meas}
\end{align}
where $\vec{\zeta}' \in \mathbb{R}^{n_d}$ is measurement noise with a covariance matrix $\mat{Z}'$.
However, such measurement cannot be directly fused using the conventional \ac{kf} framework.
The \ac{dkf} provides a method of converting this measurement model to a form that is compatible with a standard linear \ac{kf}.

Let $\mat{N}$ be a matrix such that
\begin{equation}
  \mat{N}\tran\mat{W} = \vec{0}. \label{eq:nullmat}
\end{equation}
In other words, $\mat{N} \in \mathbb{R}^{n_d\times n_z}$ is the null-space matrix of $\mat{W}$, where $n_z = n_d - n_a$.
By multiplying eq.~\eqref{eq:dkf_meas} from the left by $\mat{N}\tran$, the following relation is obtained:
\begin{equation}
  \mat{N}\tran\mat{M}\vec{x} = \mat{N}\tran\vec{o} + \mat{N}\tran\vec{\zeta}'. \label{eq:dkf_meas_tfd}
\end{equation}
It may be observed that after a substitution
\begin{align}
  \vec{z} = \mat{N}\tran\vec{o}, \hspace{0.5em} \mat{H} = \mat{N}\tran\mat{M}, \hspace{0.5em} \vec{\zeta} = -\mat{N}\tran\vec{\zeta}', \hspace{0.5em} \mat{Z} = \mat{N}\tran\mat{Z}'\mat{N}
  \label{eq:dkf_to_lkf}
\end{align}
the measurement is obtained in the standard form for a \ac{kf} defined in eq.~\eqref{eq:meas_model}.
In this manner, affine subspace measurements of a linear subspace of the state space may be fused using the standard linear \ac{kf} framework.
Note that this process is repeated with each correction step of the \ac{kf} as the variables $\mat{W}$ and $\vec{o}$, which define the \ac{dkf} measurement, change.

\subsubsection{Applying the \ac{dkf} to the Z-Axis Measurement}

In the case of the position and acceleration measurement of the \ac{ca}-\ac{kf} previously defined in this section, the \ac{dkf} measurement model from eq.~\eqref{eq:dkf_meas} corresponds to
\begin{equation}
  \bemat{
    \mat{I} & \mat{O} & \mat{O} \\
    \mat{O} & \mat{O} & \mat{I}}
  \bemat{
    \vec{p} \\
    \vec{v} \\
    \vec{a}
  }
    = \bemat{
      \vec{z}_p - \vec{\zeta}_p \\
      \frac{f_T}{m} \left( \vec{z}_{\hat{b}_3} - \vec{\zeta}_{\hat{b}_3} \right) + \vec{a}_g + \vec{a}_d
    },
    \label{eq:dkf_vertical}
\end{equation}
i.e.
\begin{align}
  \mat{M} &= \bemat{
    \mat{I} & \mat{O} & \mat{O} \\
    \mat{O} & \mat{O} & \mat{I}
  }, \hspace{0.5em}
  \mat{W} = \bemat{
    \vec{0} \\
    \vec{z}_{\hat{b}_3}
  }, \hspace{0.5em}
  \vec{\lambda} = \bemat{ \frac{f_T}{m} }, \\
  \vec{o} &= \bemat{
    \vec{z}_p \\
    \vec{a}_g
  }, \hspace{3.5em}
  \vec{\zeta}' = \bemat{
    -\vec{\zeta}_p \\
    -\frac{f_T}{m}\vec{\zeta}_{\hat{b}_3} + \vec{a}_d
  }, \\
  n_x &= 9, \hspace{2em} n_z = 5, \hspace{2em} n_d = 6, \hspace{2em} n_a = 1,
\end{align}
where $\mat{O}$ is a $3\times 3$ zero matrix.
For simplicity, we define a substitution
\begin{align}
  \vec{\zeta}_a \equiv -\frac{f_T}{m}\vec{\zeta}_{\hat{b}_3} + \vec{a}_d, && \vec{\zeta}_a \sim \mathcal{N}\left( \vec{0}, \sigma_a^2\mat{I} \right),
\end{align}
where $\vec{\zeta}_a \in \mathbb{R}^{3}$ is the collective unknown noise affecting the acceleration direction measurement, and $\sigma_a$ is the corresponding standard deviation.
This measurement model is then applied using the \ac{dkf} framework, resulting in the estimator that we refer to as the \ac{zlkf}.
Although the update law is algebraically equivalent to the formulation in~\cite{11373013}, with the \ac{dkf}, we provide a general framework not limited to this specific application.
Similarly to the position-only \acp{kf} in sec.~\ref{sec:mp_lkf}, we also evaluate both the analytical and block-diagonal process noise matrix for the \ac{zlkf}.

\section{Dataset}
\label{sec:dataset}

As the research stream explored in this paper is relatively fresh, the number of publicly available benchmarking datasets is low.
To the best of our knowledge, the only relevant dataset was published in~\cite{zhengKeypointGuidedEfficientPose2024a}, which contains RGB images of a flying \ac{uav} captured by a stationary camera in laboratory conditions with ground truth poses from motion capture.
We split this dataset based on the model of the observed \ac{uav} (Phantom~4 and Mavic~2), and use it for baseline evaluation.
However, this dataset does not provide higher-order states, which must therefore be derived retrospectively, introducing potential inaccuracies, and the motion range and agility of the target are limited (especially in the vertical direction).

To address this, we release a new dataset created using the Unreal Engine-based simulator FlightForge~\cite{capekFlightForgeAdvancingUAV2025b}, which offers photorealistic rendering and high physical fidelity with full \ac{uav} dynamics.
The dataset consists of images from a moving camera observing a target \ac{uav} following trajectories generated using the method proposed in~\cite{pliskaSafeMidAirDrone2024a} with varying levels of agility in all axes (\qtyrange{3}{21}{\metre\per\second\squared}) and at different ranges (\qtyrange{1.5}{8}{\metre}).
This allows us to evaluate the compared methods w.r.t. different flight modes and agility levels of the target.
To achieve maximal fidelity of the observed states, a full \ac{uav} control pipeline from~\cite{bacaMRSUAVSystem2021a} is used to track these trajectories.
Exact ground-truth 3D bounding boxes for detector training, full 6D pose, velocity, and acceleration are available directly from the simulator, eliminating the need for numerical differentiation of higher-order states.
Table~\ref{tab:dataset} summarizes the key characteristics of the datasets used in this work.

\begin{table}[t]
  \captionsetup{font=small, labelfont=bf, labelsep=period, skip=3pt}
  \renewcommand{\arraystretch}{1.05}
  \setlength{\tabcolsep}{2pt}
  \scriptsize
  \caption{Comparison of the benchmarking datasets.}
  \label{tab:dataset}
  \centering
  \begin{tabular}{@{}p{0.31\columnwidth}
                  p{0.21\columnwidth}
                  p{0.21\columnwidth}
                  p{0.21\columnwidth}@{}}
    \toprule
    Dataset &
    \makecell[l]{\textbf{Unreal} (ours)} &
    \makecell[l]{\textbf{Phantom 4} \cite{zhengKeypointGuidedEfficientPose2024a}} &
    \makecell[l]{\textbf{Mavic 2} \cite{zhengKeypointGuidedEfficientPose2024a}} \\
    \midrule
    Number of sequences & 140 & 68 & 61 \\
    Total duration  & \SI{1333}{\second} & \SI{667}{\second} & \SI{668}{\second} \\
    Number of frames & 33460 & 16734 & 16755 \\
    Sampling period & \SI{40}{\milli\second} & \SI{40}{\milli\second} & \SI{40}{\milli\second}   \\
    Camera motion & Flying & Static & Static \\
    Camera range$^*$  & \qtyrange{1.5}{8.0}{\metre} & \qtyrange{1.9}{5.1}{\metre} & \qtyrange{1.9}{5.1}{\metre} \\
    Ground truth source & Simulator & MoCap & MoCap \\
    UAV dimensions &
      \qtyproduct[product-units = single]{46 x 46 x 38}{\centi\metre} &
      \qtyproduct[product-units = single]{34 x 34 x 34}{\centi\metre} &
      \qtyproduct[product-units = single]{28 x 33 x 12}{\centi\metre} \\
    Velocity Q2$^\dagger$   & \SI{2.77}{\metre\per\second} & \SI{0.83}{\metre\per\second} & \SI{0.85}{\metre\per\second} \\
    Velocity p98$^\dagger$  & \SI{8.45}{\metre\per\second} & \SI{2.37}{\metre\per\second} & \SI{2.76}{\metre\per\second} \\
    Acceleration Q2$^\dagger$  & \SI{4.88}{\metre\per\second\squared}  & \SI{1.87}{\metre\per\second\squared} & \SI{1.99}{\metre\per\second\squared} \\
    Acceleration p98$^\dagger$ & \SI{15.80}{\metre\per\second\squared} & \SI{4.71}{\metre\per\second\squared} & \SI{4.67}{\metre\per\second\squared} \\
    \bottomrule
    \multicolumn{4}{l}{\footnotesize $^*$Range of the \qtyrange{2}{98}{\percent} percentile.} \\
    \multicolumn{4}{l}{\footnotesize $^\dagger$Q2, p98 denote the \SI{50}{\percent} and \SI{98}{\percent} percentiles.}
  \end{tabular}
\end{table}

\section{Results}
\label{sec:results}

\subsection{Detection and Pose Estimation}

To train the YOLOv5-6D detector~\cite{viviersAdvancing6DoFInstrument2024} without overfitting, we fine-tune it on a small random subset of the real-world datasets (\SI{2}{\percent} for the \textit{Phantom~4}, \SI{5}{\percent} for the \textit{Mavic~2} dataset).
For the \textit{Unreal} dataset, we generate an independent training set of approx. 13 thousand images.
Only the batch size is tuned as a hyperparameter, with all other settings left at their defaults.
The network is trained with heavy augmentations, including random crop, blur, and HSV shifts.
The resulting median corner reprojection error is \SI{2.73}{\pixel} on the \textit{Unreal} dataset, \SI{9.65}{\pixel} on \textit{Phantom~4}, and \SI{5.36}{\pixel} on \textit{Mavic~2}.

The next step after the 3D bounding box detection is solving the \ac{pnp} problem based on known dimensions of the target to obtain the 6D pose of the detected object.
With the above-mentioned detector, the median position error is \SI{49}{\milli\meter} on the \textit{Unreal} dataset, \SI{106}{\milli\meter} on \textit{Phantom~4}, and \SI{43}{\milli\meter} on \textit{Mavic~2}.
The median orientation error is \SI{1.6}{\degree}, \SI{3.3}{\degree}, and \SI{1.7}{\degree}, respectively.
These poses comprise the position and orientation measurements used by the evaluated estimators.

\subsection{Compared Methods}
\label{sec:setup}

We compare the proposed \ac{zlkf} estimator with the baseline \ac{cv} and \ac{ca} \acp{kf} described in sec.~\ref{sec:mp_lkf}, and with relevant state-of-the-art approaches.
Specifically, we include the Pose-KF estimator proposed in~\cite{zhengKeypointGuidedEfficientPose2024a}, which uses a \ac{cv} model and incorporates the tilt measurements through the process noise covariance matrix, and the \ac{mekf}, which is a widely adopted method for \ac{uav} ego-state estimation.
For the \ac{mekf}, we modify the standard formulation (see e.g.~\cite{mekf2015}) to correspond to the problem considered in this paper.
Specifically, we use a nominal state vector $\vec{x} = \bemat{\vec{p}\tran & \vec{v}\tran & \vec{q}\tran}\tran$, where $\vec{q}$ is a quaternion representing the \ac{uav}'s orientation, and a measurement vector $\vec{z} = \bemat{\vec{p}\tran & \vec{q}\tran}\tran + \vec{\zeta}_{pq}$, where $\vec{\zeta}_{pq}$ is the measurement noise.
We consider both the analytical and BDC formulations of the process noise covariance matrix of the \ac{mekf}, as described in sec.~\ref{sec:mp_lkf}.
Because the Pose-KF utilizes this matrix for fusion of the tilt measurement, it cannot easily be modified in the same manner, so we follow the original formulation.

\subsection{Tuning and Evaluation Protocol}
\label{sec:tune_protocol}

All individual estimator parameters are tuned using the CMA-ES~\cite{hansenReducingTimeComplexity2003} on a dedicated tuning subset for each dataset.
All reported results are computed on a disjoint held-out set of equal size, containing no samples used in the YOLOv5-6D training or validation.
For the optimization and reporting of results, we define the \ac{men}:
\begin{equation}
\text{\acs{men}} =
\frac{1}{\sum_{j=1}^{M} N_j}\sum_{j=1}^{M}\sum_{k \in \mathcal{K}_j}
  \norm{ \vec{s}_{j,k} - \hat{\vec{s}}_{j,k}  }_{2},
\label{eq:men}
\end{equation}
where $M$ is the number of test sequences, $\mathcal{K}_j$ is the set of indices of valid samples in sequence $j$, $N_j = \abs{\mathcal{K}_j}$, $\vec{s}_{j,k} \in \mathbb{R}^d$ denotes the ground-truth state vector at timestep $k$ of sequence $j$, and $\hat{\vec{s}}_{j,k}$ is the corresponding estimate.
For estimators with explicit acceleration states, $\vec{s}$ and $\hat{\vec{s}}$ include the position, velocity, and acceleration (i.e. $d=9$). For the remaining estimators, $\vec{s}$ and $\hat{\vec{s}}$ only include the position and velocity (i.e. $d=6$).
We apply a logarithmic transformation to the parameter vector during optimization $\vec{\theta} = \log_{10}\vec{\theta}_{\text{nat}}$, where $\vec{\theta}_{\text{nat}}$ is a vector of parameters used by the optimized filter, and $\vec{\theta}$ is the vector of parameters optimized by the CMA-ES algorithm.
The optimization objective also includes an additive regularization term $\lambda \sqrt{\norm{\vec{\theta}}_2}$, penalizing large deviations from the origin of the space of transformed parameters.
We empirically chose the value of the weight $\lambda = 0.001$.
These techniques proved important to stabilize convergence of the optimization by better representing parameter values spanning orders of magnitude and mitigating numerical problems.

\begin{table*}[!t]
  \captionsetup{font=small, labelfont=bf, labelsep=period, skip=3pt}
  \renewcommand{\arraystretch}{1.10}
  \setlength{\tabcolsep}{0pt}

  \caption{\ac{men} values for position, velocity, and acceleration across datasets and estimators.}
  \label{tab:men_components}
  \centering

  \begin{tabularx}{\textwidth}{@{}l
      *{3}{>{\centering\arraybackslash}X}|
      *{3}{>{\centering\arraybackslash}X}|
      *{3}{>{\centering\arraybackslash}X}@{}}
    \toprule
    & \multicolumn{3}{c|}{\textbf{Unreal} (simulated)}
    & \multicolumn{3}{c|}{\textbf{Phantom 4} (real-world)}
    & \multicolumn{3}{c}{\textbf{Mavic 2} (real-world)} \\
    \textbf{Estimator}
    & Pos. (\si{\metre}) & Vel. (\si{\metre\per\second}) & Acc. (\si{\metre\per\second\squared})
    & Pos. (\si{\metre}) & Vel. (\si{\metre\per\second}) & Acc. (\si{\metre\per\second\squared})
    & Pos. (\si{\metre}) & Vel. (\si{\metre\per\second}) & Acc. (\si{\metre\per\second\squared}) \\
    \midrule

    \multicolumn{4}{@{}l|}{\textit{Position-only}} & & & & & & \\
    CV-KF
      & 0.100 & 2.300 & --
      & 0.131 & 0.755 & --
      & \textbf{0.070} & 0.793 & -- \\
    CV-KF+BDC
      & 0.112 & \textbf{0.739} & --
      & 0.131 & 0.415 & --
      & 0.071 & \textbf{0.426} & -- \\
    CA-KF
      & 0.104 & 0.791 & \textbf{4.899}
      & 0.131 & 0.487 & 2.337
      & 0.093 & 0.609 & 2.440 \\
    CA-KF+BDC
      & \textbf{0.099} & 0.787 & 4.989
      & \textbf{0.131} & \textbf{0.415} & \textbf{2.039}
      & 0.078 & 0.488 & \textbf{2.102} \\
    \midrule

    \multicolumn{4}{@{}l|}{\textit{Pose-aware}} & & & & & & \\
    Pose-KF~\cite{zhengKeypointGuidedEfficientPose2024a}
      & 0.101 & 0.529 & --
      & 0.132 & \underline{\textbf{0.251}} & --
      & 0.072 & \underline{\textbf{0.248}} & -- \\
    MEKF~\cite{mekf2015}
      & 0.101 & 0.486 & --
      & 0.131 & 0.278 & --
      & 0.073 & 0.266 & -- \\
    MEKF+BDC
      & \underline{\textbf{0.096}} & 0.462 & --
      & \underline{\textbf{0.130}} & 0.269 & --
      & 0.068 & 0.264 & -- \\
    Z-KF
      & 0.103 & 0.431 & 2.073
      & 0.131 & 0.270 & 0.953
      & 0.073 & 0.254 & 0.876 \\
    Z-KF+BDC
      & 0.097 & \underline{\textbf{0.425}} & \underline{\textbf{2.062}}
      & 0.130 & 0.264 & \underline{\textbf{0.874}}
      & \underline{\textbf{0.067}} & 0.271 & \underline{\textbf{0.833}} \\
    \addlinespace[2pt]

    \multicolumn{4}{@{}l|}{\textit{Relative improvement (best pose-aware vs. position-only)}} & & & & & & \\
      & \phantom{00}\SI{3}{\percent} & \phantom{0}\SI{42}{\percent} & \phantom{0}\SI{58}{\percent}
      & \phantom{00}\SI{1}{\percent} & \phantom{0}\SI{40}{\percent} & \phantom{0}\SI{57}{\percent}
      & \phantom{00}\SI{4}{\percent} & \phantom{0}\SI{42}{\percent} & \phantom{0}\SI{60}{\percent} \\
    \bottomrule
  \end{tabularx}

  \vspace{3pt}\raggedright\footnotesize
  Evaluated on held-out test sequences using median-representative parameters from 10 independent tuning runs.
  Acceleration is reported only for estimators with explicit acceleration states.
  \textbf{Bold}: best in category. \underline{Underline}: best overall.
  \vspace{-5mm}
\end{table*}

Each estimator was independently tuned ten times.
All reported results correspond to the median run.
It is worth noting that the final loss was extremely consistent between runs for each filter (within \SI{1}{\percent} between the worst and best results), except for the BDC variants, where the difference in the loss was up to \SI{175}{\percent}, which we attribute to the increased number of optimized parameters.
Table~\ref{tab:men_components} summarizes the median-run \ac{men} across all estimators and datasets.
To verify that using a different optimization metric does not bias the results, we also tried tuning all estimators with the same metric, using only position and velocity ($d=6$).
The relative ranking of the estimators was preserved, and the comparable metrics changed only negligibly ($< \SI{1}{\percent}$).
However, including the acceleration in the loss function for filters that support it significantly improves the acceleration estimation, and thus also the prediction, which is crucial for multi-robot interaction.

\subsection{Estimation Accuracy}
\label{sec:accuracy}

Table~\ref{tab:men_components} reports position, velocity, and acceleration \ac{men} across all estimators and datasets.
Position \ac{men} is approximately constant across all methods (${\sim}\SI{0.10}{\metre}$), showing that position accuracy is primarily determined by the accuracy of the detector and \ac{pnp} solver, not the filter structure.

The \ac{cv}-\ac{kf} exhibits the highest velocity error across all datasets, confirming that a sufficient order of the model is a prerequisite for tracking maneuvering targets.
Its BDC variant reduces the velocity error by \SI{53}{\percent} on average, exceeding even the raw \ac{ca}-\ac{kf}, which demonstrates that an expressive process-noise model can effectively compensate model mismatch.
For the \ac{ca}-\ac{kf}, the improvement of using BDC over analytical process noise is less dramatic (\SI{12}{\percent} on average), indicating that once model order is adequate, the primary bottleneck is the absence of thrust-direction information rather than process-noise modeling.

Pose-KF~\cite{zhengKeypointGuidedEfficientPose2024a} achieves the lowest velocity \ac{men} on both real-world datasets, outperforming \ac{zlkf}+BDC by \SI{5}{\percent} on \textit{Phantom~4} and \SI{8}{\percent} on \textit{Mavic~2}.
However, it does not estimate acceleration, and its process model assumes constant thrust magnitude, biasing the filter to favor less agile maneuvers constrained to one horizontal plane.
On the challenging \textit{Unreal} dataset, which includes agile motion spanning all three axes equally, \ac{zlkf}+BDC outperforms Pose-KF by \SI{20}{\percent} in velocity.

\ac{mekf}~\cite{mekf2015} falls between Pose-KF and \ac{zlkf}, confirming that thrust-direction information improves estimation regardless of the specific filter formulation.
\ac{zlkf}+BDC achieves the overall lowest velocity and acceleration errors on the \textit{Unreal} dataset, reducing \ac{men} by \SI{42}{\percent} and \SI{58}{\percent} relative to the best position-only variant.
The difference between \ac{zlkf} and \ac{zlkf}+BDC is marginal (${\sim}\SI{1}{\percent}$), showing the same pattern as \ac{ca}-\ac{kf}: once the dominant information source is present, process-noise refinement yields diminishing returns.

\subsection{Performance over Agility}

\begin{figure}[t]
\centering
\captionsetup{skip=6pt, labelfont=bf, labelsep=period, font=small}
\includegraphics[width=\columnwidth]{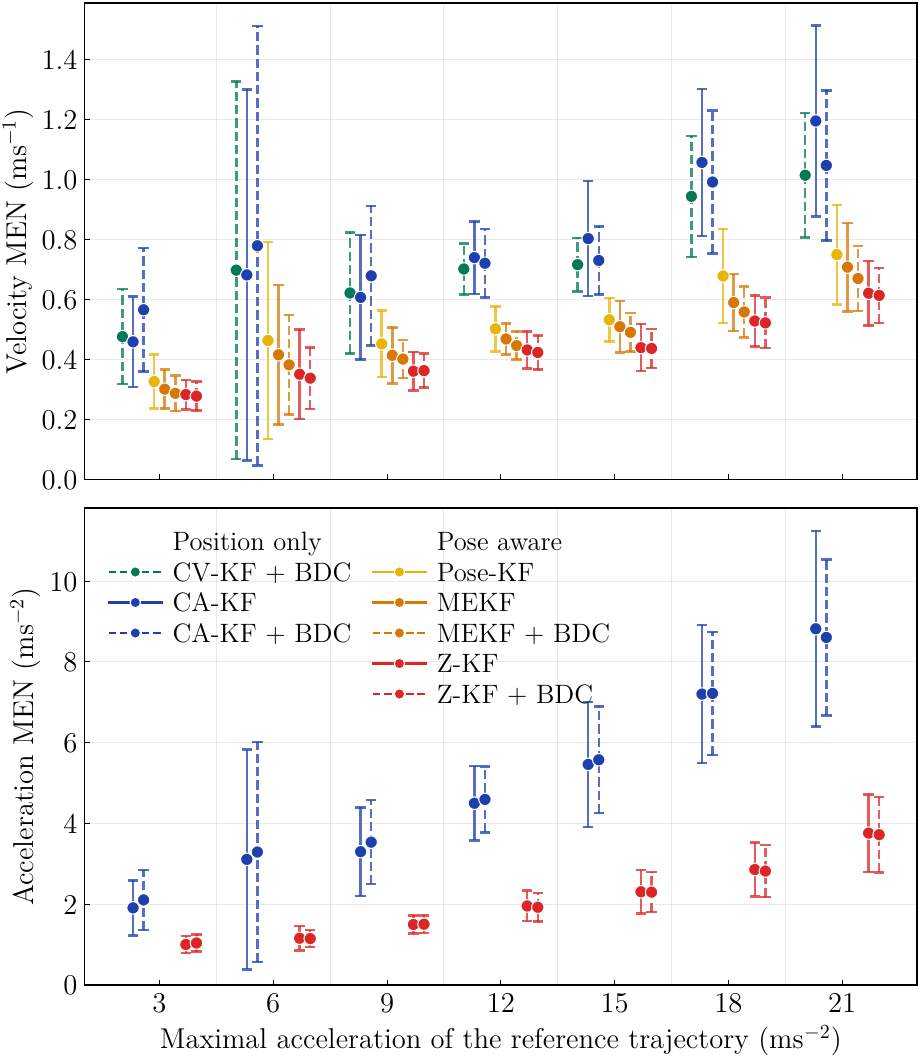}
  \caption{Velocity and acceleration \ac{men} for the different agility levels in the \textit{Unreal} dataset. \ac{cv}-\ac{kf} is omitted due to a significantly higher error at all levels (see Table~\ref{tab:men_components}). Dots denote mean over 10 trajectories, lines indicate standard deviation.}
\label{fig:agility_scaling}
\end{figure}

\begin{table}
  \captionsetup{font=small, labelfont=bf, labelsep=period, skip=3pt}
  \caption{Agility levels in the \textit{Unreal} (simulated) dataset.}
  \label{tab:agility_levels}
  \centering
  \renewcommand{\arraystretch}{1.10}
  \setlength{\tabcolsep}{5.0pt}
  \begin{tabularx}{\columnwidth}{llll|ll|ll}
    \toprule
      Agility group
      & \multicolumn{3}{c|}{Low}
      & \multicolumn{2}{c|}{Mid}
      & \multicolumn{2}{c}{High} \\
      Reference max. (\si{\metre\per\second\squared})
      & 3.0 & 6.0 & 9.0
      & 12.0 & 15.0
      & 18.0 & 21.0
      \\
      Flown acc. Q2$^\dagger$ (\si{\metre\per\second\squared})
      & 1.6 & 3.0 & 4.0
      & \phantom{1}5.7 & \phantom{1}7.0
      & \phantom{1}7.9 & 10.0
      \\
      Flown acc. p98$^\dagger$ (\si{\metre\per\second\squared})
      & 2.8 & 5.7 & 7.9
      & 11.1 & 14.0
      & 15.9 & 19.1
      \\
      Flown tilt p98$^\dagger$ (\si{\degree})
      & 15 & 32 & 45
      & 62 & 71
      & 86 & 95
      \\
      Num. of sequences
      & 20 & 20 & 20
      & 20 & 20
      & 20 & 20
      \\
    \bottomrule
    \multicolumn{8}{l}{\footnotesize $^\dagger$Q2, p98 denote the \SI{50}{\percent} and \SI{98}{\percent} percentiles.}
  \end{tabularx}
\end{table}

Our \textit{Unreal} dataset is divided into seven agility levels based on the maximal acceleration of the generated reference trajectory (see Tab.~\ref{tab:agility_levels}).
Figure~\ref{fig:agility_scaling} shows velocity and acceleration \ac{men} for each estimator across these levels.
Pose-aware filters outperform position-only variants at every agility level with a consistently smaller estimation error and error variance.
Both groups exhibit an increasing error with agility, but position-only errors grow faster, widening the absolute gap.
\ac{zlkf} achieves the lowest error and variance throughout.
The acceleration plot shows an even clearer separation: \ac{zlkf} maintains a low error with remarkably tight variance at all agility levels, while \ac{ca}-\ac{kf} error and variance both grow rapidly.
This indicates that thrust-direction information improves estimation not only during agile flight, but in all operating conditions.

\begin{table}[t]
    \captionsetup{font=small, labelfont=bf, labelsep=period, skip=3pt}
    \centering
    \caption{Per-axis \ac{men} values on the \textit{Unreal} dataset across the different agility groups.}
    \label{tab:per_axis_men}
    \renewcommand{\arraystretch}{1.10}
    \setlength{\tabcolsep}{6.5pt}
    
    \begin{tabularx}{\columnwidth}{Xcccccc}
    \toprule
    & \multicolumn{2}{c}{\textbf{Low} (3--9)}
    & \multicolumn{2}{c}{\textbf{Mid} (12--15)}
    & \multicolumn{2}{c}{\textbf{High} (18--21)} \\
    \cmidrule{2-7}
    \textbf{Estimator}
    & $xy$ & $z$ & $xy$ & $z$ & $xy$ & $z$ \\
    \midrule
    
    \multicolumn{7}{@{}l@{}}{\textit{Velocity \ac{men} (\si{\metre\per\second})}} \\
    CA-KF+BDC  & 0.376 & 0.248 & 0.393 & 0.306 & 0.553 & 0.404 \\
    Pose-KF~\cite{zhengKeypointGuidedEfficientPose2024a}   & 0.194 & 0.220 & 0.219 & 0.332 & 0.312 & 0.444 \\
    Z-KF+BDC   & \textbf{0.136} & \textbf{0.207} & \textbf{0.192} & \textbf{0.260} & \textbf{0.259} & \textbf{0.330} \\
    \multicolumn{7}{@{}l@{}}{\textit{Relative improvement of Z-KF+BDC over CA-KF+BDC}} \\
      & \SI{64}{\percent} & \SI{17}{\percent} & \SI{51}{\percent} & \SI{15}{\percent} & \SI{53}{\percent} & \SI{18}{\percent} \\
    \midrule
    
    \multicolumn{7}{@{}l@{}}{\textit{Acceleration \ac{men} (\si{\metre\per\second\squared})}} \\
    CA-KF+BDC  & 1.646 & 1.105 & 2.793 & 1.889 & 4.385 & 2.885 \\
    Z-KF+BDC   & \textbf{0.314} & \textbf{1.054} & \textbf{0.768} & \textbf{1.548} & \textbf{1.382} & \textbf{2.135} \\
    \multicolumn{7}{@{}l@{}}{\textit{Relative improvement of Z-KF+BDC over CA-KF+BDC}} \\
      & \SI{81}{\percent} & \SI{5}{\percent} & \SI{73}{\percent} & \SI{18}{\percent} & \SI{68}{\percent} & \SI{26}{\percent} \\
    \bottomrule
    \end{tabularx}
\end{table}

To understand the source of the improvement, Table~\ref{tab:per_axis_men} decomposes \ac{men} into horizontal ($xy$) and vertical ($z$) components.
The improvement in velocity estimation of \ac{zlkf}+BDC over \ac{ca}-\ac{kf}+BDC is significantly higher in the horizontal direction than in the vertical.
This asymmetry confirms that thrust-direction sensing primarily encodes lateral acceleration.
However, the consistent \qtyrange{15}{18}{\percent} vertical improvement is also non-negligible.
For acceleration, the contrast is sharper (especially at low agility), indicating that the large reduction in acceleration error originates specifically from thrust-direction observation of lateral motion.
Pose-KF $z$-axis velocity error exceeds even \ac{ca}-\ac{kf}+BDC at mid and high agility levels, highlighting how its constant thrust-magnitude assumption degrades vertical estimation under aggressive tilt.

\subsection{Estimation Latency}
\label{sec:latency}
\begin{figure}[t]
\centering
\captionsetup{skip=6pt, labelfont=bf, labelsep=period, font=small}
 \includegraphics[width=\columnwidth]{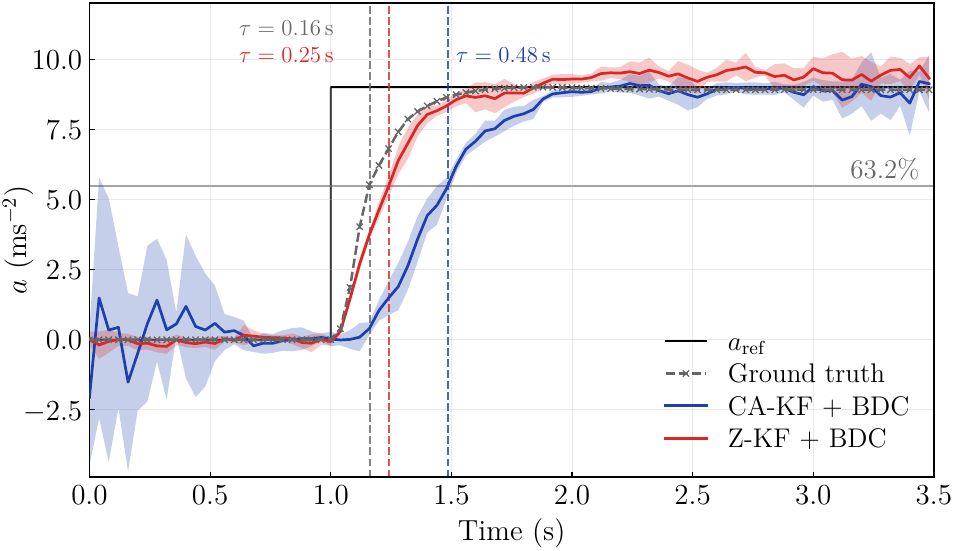}
\caption{Acceleration step response at \SI{9}{\metre\per\second\squared} magnitude. Shaded regions indicate variance across five runs. Crosses represent camera frames. %
}
\label{fig:step_response}
\end{figure}

Beyond estimation accuracy, agile multi-\ac{uav} flight demands low-latency state estimates.
For example, in a leader-follower scenario, the estimation delay directly determines the ability to keep the leader within the camera's field of view.
To quantify this, we design a dedicated experiment in which the observed \ac{uav} executes acceleration step commands in an arbitrary horizontal direction at each of the seven agility levels.
Five camera trajectories are applied to each recording to enable a fair comparison across viewpoints.
We compare the best-performing position-only (\ac{ca}-\ac{kf}+BDC) and pose-aware (\ac{zlkf}+BDC) estimators.
Figure~\ref{fig:step_response} shows a representative step response at $\SI{9}{\metre\per\second\squared}$.
The reaction time of the \ac{ca}-\ac{kf}+BDC w.r.t. the system's response is visibly delayed, while the \ac{zlkf}+BDC responds near-instantly, as the thrust-direction measurement captures the acceleration onset before significant displacement accumulates.

Table~\ref{tab:latency} quantifies this across all agility levels using the classical \SI{63.2}{\percent} rise time.
We compare the ground-truth rise time $\tau_{\text{gt}}$, determined by the \ac{uav}'s dynamics, with the rise time of the estimated acceleration.
\ac{ca}-\ac{kf}+BDC introduces an approximately constant overhead of ${\sim}\SI{300}{\milli\second}$ (corresponding to ${\sim}7$--$8$ frames) above the physical onset, independent of agility level, which may be counterintuitive.
A higher acceleration produces proportionally larger inter-frame displacements, yet the detection delay does not decrease.
This points to a structural bottleneck in position-based estimation, where a fixed number of displacement samples is required to infer acceleration regardless of its magnitude.
In contrast, \ac{zlkf}+BDC operates near the physical detection limit at low agility.
At higher agility, its delay gradually increases, but remains below the \ac{ca}-\ac{kf}+BDC at every level.

\begin{table}[t]
\captionsetup{font=small, labelfont=bf, labelsep=period, skip=3pt}
\centering
\caption{Step response delay over reference magnitude.}
\label{tab:latency}
\renewcommand{\arraystretch}{1.15}
\begin{tabular*}{\columnwidth}{@{\extracolsep{\fill}} r | rr | rr | rr}
\toprule
$a_{\mathrm{ref}}$
& \multicolumn{2}{c|}{$\tau_{\text{gt}}$}
& \multicolumn{2}{c|}{$\tau_{\text{CA-KF+BDC}}$}
& \multicolumn{2}{c}{$\tau_{\text{Z-KF+BDC}}$} \\
(\si{\metre\per\second\squared})
& \si{\milli\second} & frames
& \si{\milli\second} & frames
& \si{\milli\second} & frames \\
\midrule
3  & 152.0 & 3.8 & 456.4 & 11.4 & 156.0 & 3.9 \\
6  & 159.2 & 4.0 & 482.9 & 12.1 & 208.4 & 5.2 \\
9  & 164.4 & 4.1 & 477.6 & 11.9 & 247.3 & 6.2 \\
12 & 172.1 & 4.3 & 474.8 & 11.9 & 292.1 & 7.3 \\
15 & 180.6 & 4.5 & 476.1 & 11.9 & 327.0 & 8.2 \\
18 & 187.9 & 4.7 & 482.8 & 12.1 & 344.6 & 8.6 \\
21 & 170.2 & 4.3 & 460.7 & 11.5 & 336.6 & 8.4 \\
\bottomrule
\end{tabular*}

  \vspace{1mm}
  \raggedright\footnotesize
  Response delays of the physical system ($\tau_{\text{gt}}$) and of the best position-only and pose-aware estimators. The delays are reported as the \SI{63.2}{\percent} rise time. Values are averaged over five samples (std.\ dev.\ $< \SI{34}{\milli\second}$ for all
conditions).
\end{table}

\subsection{Closed-Loop Validation}
\label{sec:closedloop}

To demonstrate the system-level impact of estimation quality, a leader-follower scenario is evaluated in a closed-loop experiment using a high-fidelity photorealistic simulator~\cite{capekFlightForgeAdvancingUAV2025b}.
The leader \ac{uav} flies along the $y$-axis between increasingly distant waypoints, reaching accelerations above \SI{20}{\metre\per\second\squared}.
The follower, offset by \SI{4}{\metre} in $x$, observes the leader through a \SI{110}{\degree} FOV camera using the full perception pipeline (YOLOv5-6D, \ac{pnp}, state estimator).
The estimated state is propagated over a \SI{1.4}{\second} horizon, matching the NMPC control horizon~\cite{guptaLoLNMPCLowLevelDynamics2025}.
The estimator parameters are re-tuned on \ac{men} error over the predicted trajectory, aligning the objective with the control task.
We compare the \ac{zlkf} and \ac{ca}-\ac{kf}.

When the \ac{ca}-\ac{kf} is employed, the follower becomes unstable already during hovering, losing the leader within approximately \SI{5}{\second}.
Estimation errors compound over the \SI{1.4}{\second} prediction horizon, causing progressively larger control corrections until the leader exits the camera field of view.
Both variants (\ac{ca}-\ac{kf}, \ac{ca}-\ac{kf}+BDC) exhibit the same failure.

\begin{figure}[t]
\centering
\captionsetup{skip=6pt, labelfont=bf, labelsep=period, font=small}
\includegraphics[width=\columnwidth]{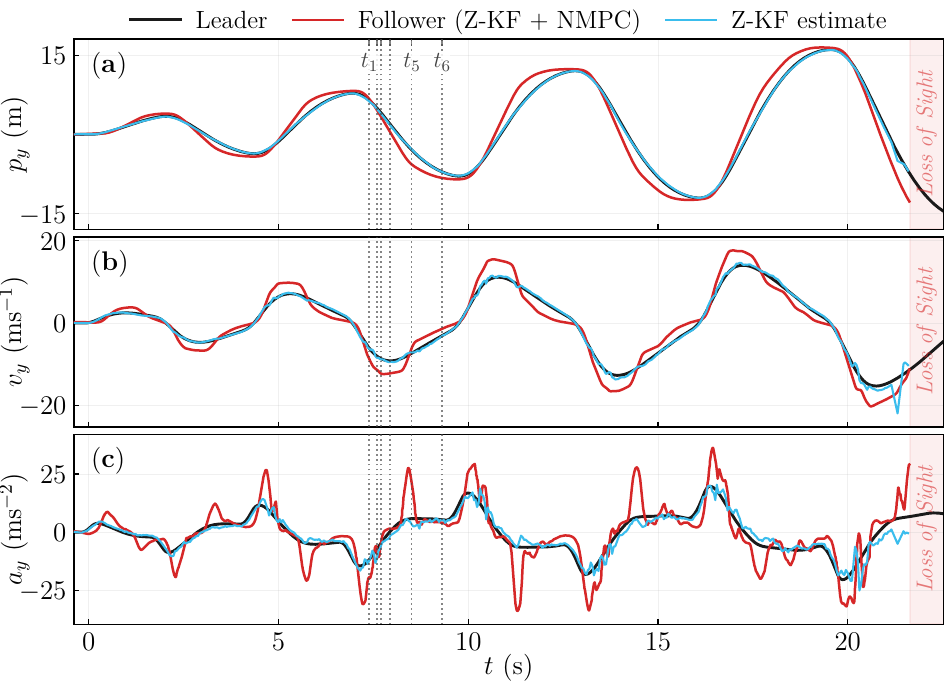}
  \caption{Closed-loop leader-follower tracking with \ac{zlkf} and NMPC.
Note that with a position-only estimator, the follower is not even capable of stable hovering (not shown).
Markers $t_1$--$t_6$ reference Fig.~\ref{fig:sim_timelapse}.
(a)~Position. (b)~Velocity. (c)~Acceleration.}
\label{fig:closedloop}
\end{figure}

With \ac{zlkf}, the follower maintains stable hovering and subsequently tracks the leader through lateral maneuvers above \SI{2}{\gee} (up to \SI{60}{\degree} tilt, \SI{14}{\metre\per\second}) over a \SI{20}{\second} sequence (see Fig.~\ref{fig:closedloop}).
The follower itself reaches accelerations above $\SI{25}{\metre\per\second\squared}$ due to corrective control.
The experiment ends when the leader exits the camera's field of view at the most aggressive waypoint, not due to a failure of the estimator.

\section{Discussion \& Conclusion}

This work provided a detailed analysis of the influence of including thrust-direction information in vision-based state estimation of multirotor \acp{uav} over using only position measurements.
The main limitation of position-only estimation is structural: higher-order states must be inferred from accumulated displacement.
This problem manifests already at low agility levels, substantially affecting estimation accuracy.
Moreover, it introduces an inherent estimation lag, bounding the achievable agility when used in closed-loop control.
Incorporating the thrust direction significantly reduces the estimation error and delay.

Although we used real-world datasets and high-fidelity simulation to ensure robust evaluation, several limitations remain.
Our \ac{pnp}-based pose extraction assumes known dimensions of the target, which may be limiting in some scenarios, but this can be solved e.g. by utilizing a stereo camera for distance measurement.
The closed-loop experiments were conducted only in simulation, with real-world validation remaining an open problem, which we plan to address in future work.

Overall, the presented results demonstrate that to achieve truly agile communication-free multi-\ac{uav} flight, conventional state estimation approaches based on relative localization are insufficient.
We provide a complete methodology to remove this limitation that we believe will be fundamental in further research in agile multi-\ac{uav} flight.

\bibliographystyle{IEEEtran}
\bibliography{MRS}

\end{document}

%% file: acronyms.tex
\acrodef{AAIS}[AAIS]{Autonomous Aerial Interception System}
\acrodefplural{AAIS}[AAIS']{Autonomous Aerial Interception Systems}

\acrodef{C-UAS}[C-UAS]{Counter Unmanned Aircraft System}
\acrodefplural{C-UAS}[C-UAS']{Counter Unmanned Aircraft Systems}

\acrodef{FoV}[FoV]{Field of View}
\acrodef{VFoV}[VFoV]{Vertical Field of View}
\acrodef{HFoV}[HFoV]{Horizontal Field of View}
\acrodef{RL}[RL]{Reinforcement Learning}
\acrodef{ARL}[ARL]{Application Readiness Level}
\acrodef{BFS}[BFS]{Breadth-First Search}
\acrodef{GPS}[GPS]{Global Positioning System}
\acrodef{SLAM}[SLAM]{Simultaneous Localization And Mapping}
\acrodef{SLAMs}[SLAMs]{Simultaneous Localization And Mapping systems}
\acrodef{GPS}[GPS]{Global Positioning System}
\acrodef{RTK}[RTK]{Real-time Kinematic}
\acrodef{GNSS}[GNSS]{Global Navigation Satellite System}
\acrodef{ROS}[ROS]{Robot Operating System}
\acrodef{API}[API]{Application Programming Interface}
\acrodef{UGV}[UGV]{Unmanned Ground Vehicle}
\acrodef{UV}[UV]{Ultra-Violet}
\acrodef{LED}[LED]{Light-emitting Diode}
\acrodef{MBZIRC}[MBZIRC]{Mohamed Bin Zayed International Robotics Challenge}
\acrodef{DARPA}[DARPA]{Defense Advanced Research Projects Agency}
\acrodef{SAR}[SAR]{Search and Rescue}
\acrodef{IMU}[IMU]{Inertial Measurement Unit}
\acrodef{LTI}[LTI]{Linear time-invariant}
\acrodef{MPC}[MPC]{Model Predictive Control}
\acrodef{UVDAR}[UVDAR]{Ultra-Violet Direction And Ranging}
\acrodef{DOF}[DOF]{degree-of-freedom}
\acrodef{DOFs}[DOFs]{degrees-of-freedom}
\acrodef{LiDAR}[LiDAR]{Light Detection and Ranging}
\acrodef{ESC}[ESC]{Electronic Speed Controller}
\acrodef{ICP}[ICP]{Iterative Closest Points}
\acrodef{kf}[KF]{Kalman Filter}
\acrodef{lkf}[LKF]{Linear Kalman Filter}
\acrodef{plkf}[PLKF]{Pseudo-Linear Kalman Filter}
\acrodef{ukf}[UKF]{Unscented Kalman Filter}
\acrodef{ekf}[EKF]{Extended Kalman Filter}
\acrodef{dkf}[DKF]{Degenerate Kalman Filter}
\acrodef{RAS}[RAS]{Robotics and Automation Society}
\acrodef{IEEE}[IEEE]{Institute of Electrical and Electronics Engineers}
\acrodef{MRS}[MRS]{Multi-robot Systems Group}
\acrodef{CNN}[CNN]{Convolutional Neural Network}
\acrodef{CTU}[CTU]{Czech Technical University}
\acrodef{UPenn}[UPenn]{University of Pennsylvania}
\acrodef{NYU}[NYU]{New York University}
\acrodef{FIFO}[FIFO]{First In, First Out}
\acrodef{RMSE}[RMSE]{Root Mean Square Error}
\acrodef{PDF}[PDF]{Probability Distribution Function}
\acrodef{CDF}[CDF]{Cumulative Distribution Function}
\acrodef{MC}[MC]{Monte-Carlo}
\acrodef{TSDF}[TSDF]{Truncated Signed Distance Field}
\acrodef{SITL}[SITL]{Software In The Loop}

\acrodef{PP}[PP]{Pure Pursuit}
\acrodef{PN}[PN]{Proportional Navigation}
\acrodef{PPN}[PPN]{Pure Proportional Navigation}
\acrodef{TPN}[TPN]{True Proportional Navigation}
\acrodef{LPN}[LPN]{Linearized Proportional Navigation}

\acrodef{FOV}[FOV]{Field Of View}
\acrodef{LOS}[LOS]{Line Of Sight}
\acrodef{ZEM}[ZEM]{Zero-Effort Miss}
\acrodef{SWaP}[SWaP]{Size, Weight, and Power}

\acrodef{EPN}[FRPN]{Fast Response Proportional Navigation}
\acrodef{GPN}[GPN]{General Proportional Navigation}
\acrodef{VoFOD}[VoFOD]{Volumetric Flying Object Detector}

\acrodef{aais}[AAIS]{Autonomous Aerial Interception System}
\acrodef{arl}[ARL]{Application Readiness Level}
\acrodef{ca}[CA]{Constant Acceleration}
\acrodef{cv}[CV]{Constant Velocity}
\acrodef{ekf}[EKF]{Extended Kalman Filter}
\acrodef{frpn}[FRPN]{Fast Response Proportional Navigation}
\acrodef{gpn}[GPN]{Generalized Proportional Navigation}
\acrodef{gnss}[GNSS]{Global Navigation Satellite System}
\acrodef{ibvs}[IBVS]{Image-Based Visual Servoing}
\acrodef{imm}[IMM]{Interacting Multiple Model}
\acrodef{irl}[IRL]{Inverse Reinforcement Learning}
\acrodef{kf}[KF]{Kalman Filter}
\acrodef{los}[LOS]{Line of Sight}
\acrodef{mpc}[MPC]{Model Predictive Control}
\acrodef{marl}[MARL]{Multi-Agent Reinforcement Learning}
\acrodef{nmpc}[NMPC]{Nonlinear Model Predictive Control}
\acrodef{pca}[PCA]{Principal Component Analysis}
\acrodef{pn}[PN]{Proportional Navigation}
\acrodef{rf}[RF]{Radio Frequency}
\acrodef{rl}[RL]{Reinforcement Learning}
\acrodef{sota}[SotA]{State of the Art}
\acrodef{vofod}[VoFOD]{Volumetric Flying Object Detector}
\acrodef{uae5}[UAE5]{Unreal Engine 5}
\acrodef{uav}[UAV]{Unmanned Aerial Vehicle}
\acrodef{rmse}[RMSE]{Root Mean Square Error}
\acrodef{pnp}[PnP]{Perspective-n-Point}
\acrodef{zlkf}[Z-KF]{Z-axis measurement Kalman Filter}
\acrodef{mekf}[MEKF]{Multiplicative Extended Kalman Filter}
\acrodef{men}[MEN]{Mean Error Norm}

%% file: common.tex
\newcommand{\norm}[1]{\left\lVert#1\right\rVert}
\newcommand{\abs}[1]{\left\lvert#1\right\rvert}

\usepackage[e]{esvect}

\renewcommand{\vec}[1]{\bm{#1}}

\newcommand{\mat}[1]{\mathbf{#1}}

\newcommand{\bemat}[1]
{
  \begin{bmatrix}
    #1
  \end{bmatrix}
}

\newcommand{\tstep}[2][]{_{#1[#2]}}
\newcommand*{\tran}{^{\intercal}}

\DeclareSIUnit{\pixel}{px}
\DeclareSIUnit{\fps}{FPS}
\DeclareSIUnit{\gee}{g}

\let\originalleft\left
\let\originalright\right
\renewcommand{\left}{\mathopen{}\mathclose\bgroup\originalleft}
\renewcommand{\right}{\aftergroup\egroup\originalright}

\newcommand{\set}[1]{\mathcal{\expandafter\MakeUppercase\expandafter{#1}}}

\newcommand{\at}[2]{\left.\kern-\nulldelimiterspace#1\right|_{#2}}